# From AI-Enabled Remote Patient Monitoring Toward Agentic Robotics: Embodiment, Governance, and Care Integration

Ivana Ragusa,[1,2] Lorenzo Guiducci,[1] and Filippo Cavallo[1,2]

[1]Department of Industrial Engineering, University of Florence, Florence, Italy; email: filippo.cavallo@unifi.it

[2]CoAImed S.r.l., Sesto Fiorentino, Florence, Italy

**Keywords**

remote patient monitoring, RPM, artificial intelligence, agentic AI, embodied AI, healthcare robotics, AI governance, digital health

**Abstract**

AI-enabled remote patient monitoring is evolving beyond passive data collection toward agentic AI systems that interpret patient states, coordinate care, and initiate action across domestic and clinical settings. This structured scoping review finds that mature agentic care has not yet emerged: Most systems remain human led, while stronger decision–action coupling and embodied AI remain largely confined to prototypes and pilots. Healthcare robotics makes the transition from monitoring to action especially visible by translating AI-generated recommendations and decisions into speech, movement, prompting, physical assistance, escalation, and care coordination, making embodiment, clinical integration, human oversight, and governance fundamental design requirements. Further progress requires safe and interoperable care architectures in which intelligent components operate through clinically validated and accountable workflows. We propose agentic robotics as a future direction for supervised systems embedded in the care ecosystem, connecting AI agents and robots with clinicians, caregivers, and patients through auditable workflows that can be interrupted and remain subject to human oversight and explicit AI governance.

## 1. INTRODUCTION

Global aging and the expansion of care beyond institutional settings are creating sustained demand for technologies that can support independent living while reducing the burden on clinicians, caregivers, and families (1). Over the last few decades, remote care technologies have progressively evolved from domotics (referring to the use of automated and networked technologies to monitor and control functions within the home) and early smart environments (which focused mainly on environmental control, safety, and remote monitoring) toward more integrated ecosystems built on the Internet of Things (IoT), wearable sensors, ambient intelligence, and AI (2). These systems have expanded the possibilities of home-based care by enabling continuous sensing, context awareness, data-driven prediction, and more personalized forms of assistance. However, monitoring alone is often insufficient, since in a caring context, sensing must be also translated into safe, meaningful, and timely support within complex domestic environments.

Robotic technologies represent a further step in this evolution. By providing AI-based services with physical and social embodiment, robots can make digital intelligence concrete and actionable in everyday life (3). In domestic environment, they can support daily routines through navigation and object manipulation. Their verbal and nonverbal communication capabilities also allow them to provide companionship and mediate care interactions among users, caregivers, and clinicians. Together, these functions support a broad range of use cases and scenarios, from physical assistance with feeding, dressing, mobility and household activities to cognitive and social support, health monitoring, medication management and emergency detection (4). Beyond the diversity of applications, robotics for home-based care poses distinctive control challenges. Autonomous systems must adapt their behavior to changes in users' needs and conditions under partial observability, while coordinating with multiple care actors and operating within strict safety and ethical constraints. These demands make remote care a particularly rich testbed for integrating perception and planning with safe control and effective human–robot interaction.

In particular, personalized care robots designed for older adults and other vulnerable populations, including people living with frailty, cognitive or physical impairments, or neurodegenerative conditions, illustrate how assistive technologies can operate in home and community settings, and adapt to individual needs, cognitive profiles, preferences, and changing conditions while remaining acceptable in real-world domestic use (5). In this context, autonomous systems must implement a full perception–inference–decision–action loop under strict safety and reliability requirements. Remote care is one of the most demanding instantiations of this problem because it addresses adverse events such as falls, nonadherence medication, dementia and cognitive decline, chronic disease progression, mental health crisis, and social isolation. These situations unfold in homes where sensory and contextual information is often incomplete and responsibility is distributed across patients, caregivers, clinicians, and devices.

With the maturation of IoT and AI, the focus progressively shifted toward more complex care architectures that integrate data through interoperable and standardized systems, enabling context awareness and decision support. More recently, advances in robotic technologies have introduced the Internet of Robotic Things (IoRT) paradigm (6), which connects robots with distributed sensing and AI-enabled infrastructures allowing monitoring to lead to situated interaction and action (7). This evolution is further reinforced by emerging agentic AI approaches, which increase the potential for autonomous planning, adaptation, and coordination across home care services (8).

From this perspective, robotics becomes a key enabling technology when AI-enabled remote healthcare must move beyond digital inference to act directly in the physical environment. Robots provide AI with an embodied presence, enabling sensing and reasoning to guide situated acting in domestic environments. This embodiment creates technical, architectural and governance challenges for remote setting, since safe and meaningful deployment depends on interoperable and accountable systems that are clinically relevant, and acceptable to users, caregivers, and healthcare organizations. This article therefore examines how robotic embodiment is incorporated into AI-enabled remote patient monitoring (RPM), ambient assisted living (AAL) and remote or distributed care architectures, with particular attention to how architecture and governance shape the transition from monitoring to real-world care action.

Robotic prototypes were included in the core corpus only when embedded in, or explicitly specified as part of, an architecture connecting sensing and inference with monitoring, decision support, actuation, escalation, or governance. Accordingly, the observed prevalence of embodiment refers in particular to architecturally integrated robotics within AI-enabled RPM/AAL corpus and should not be generalized to healthcare robotics as a whole. Complete core and background corpus registries, including the embodied robotic subset, are provided in **Supplemental Tables 19–21**.

Addressing the growing interest in agentic AI for healthcare , this structured scoping review, treating agency as a system-level property of the care architecture, maps how sensing, inference, decision, actuation, and governance (SIDAG) are configured across the systems included in the corpus and identifies which configurations incorporate robotic embodiment. The study also examines whether embodiment is associated with stronger decision–action coupling and identifies barriers to sustained deployment related to system maturity, governance, and ecosystem integration. The review provides an evidence map of concrete system architectures, while clinical effectiveness lies out of its scope.

## 2. METHODOLOGY

### 2.1. Review Aims and Evidence-Mapping Design

The Preferred Reporting Items for Systematic Reviews and Meta-Analyses Extension for Scoping Reviews (PRISMA-ScR) (9) was used as transparency guidance for reporting search, source selection, and evidence charting, while the corpus itself was used to support conceptual synthesis rather than statistical meta-analysis. We therefore prioritized papers that described implemented or clearly specified systems over papers that made only abstract claims about autonomy, robotics, agentic AI, or digital health. The review uses the broader RPM/AAL corpus as a baseline and robotic embodiment as an analytical lens for identifying when monitoring begins to become situated care action. The review was organized around three evidence-mapping questions:

- EQ1: How are AI-enabled RPM, AAL, and robotic care systems architecturally configured, and where do they move beyond monitoring toward care action?
- EQ2: How are embodied robotic systems positioned across decision–action coupling and maturity levels compared with nonembodied RPM systems?
- EQ3: Which architectural patterns characterize action-capable systems with real-user validation, and what systematic gaps prevent wider real-world deployment?

These questions were operationalized in the search strategy through four query blocks, summarized in **Table 1**.

Table 1 Query blocks used to construct the evidence-mapping corpus

| Query block | Primary evidence-mapping questions | Functions |
|---|---|---|
| Block 1: domain context | EQ1–EQ3 | RPM/AAL, telecare, home monitoring, robotic assistance |
| Block 2: decision–action coupling | EQ2 and EQ3 | Alerts, recommendations, triage, coordination, device or robot action |
| Block 3: architecture/interoperability | EQ1 and EQ3 | Architecture, interoperability, edge/cloud, supervision, safety, governance |
| Block 4: exclusion terms | EQ1–EQ3 | Iteratively identified non-health noise |

Abbreviations: AAL, ambient assisted living; RPM, remote patient monitoring.

Searches were conducted in Scopus, PubMed/MEDLINE, and IEEE Xplore using three positive query blocks covering remote care and assistive robotics domains, decision–action coupling, and architecture/interoperability. A fourth block excluded recurrently retrieved records from clearly unrelated domains, such as industrial automation, agriculture, home security, autonomous vehicles, warehouses, airports, and surgical robotics. Searches were limited to English-language peer-reviewed journal articles and conference papers published from 2018 to 2026. The searches identified 516 records (317 from Scopus, 113 from PubMed/MEDLINE, and 86 from IEEE Xplore). After duplicate removal, 448 unique records entered screening. Database-level yields, deduplication procedures, and the complete source selection flow are reported in **Supplemental Tables 1**, **2**, and **6**.

Screening and eligibility assessment were performed independently and blinded by two reviewers (I.R. and L.G.) using prespecified inclusion and exclusion criteria. Full texts were consulted whenever eligibility could not be determined from the title and abstract alone. Uncertain or discordant cases were resolved through consensus under the supervision of a third reviewer (F.C.). Preconsensus agreement was substantial (Cohen's $\kappa = 0.82$). Reviewer-specific screening outcomes, the preconsensus contingency table, agreement statistics, and consensus resolution details are provided in **Supplemental Tables 7–11**.

The final evidence-mapping corpus included 113 records: 93 core system papers used for corpus-level synthesis and 20 background papers used for conceptual, regulatory, clinical, and robotics framing. The source selection process is summarized in **Figure 1**. Sections within the **Supplemental Material** provides complete search strings (Section 4), eligibility criteria (Section 5), deduplication records (Section 6.1), screening agreement (Section 6.3), and source selection tables (Section 6.5).

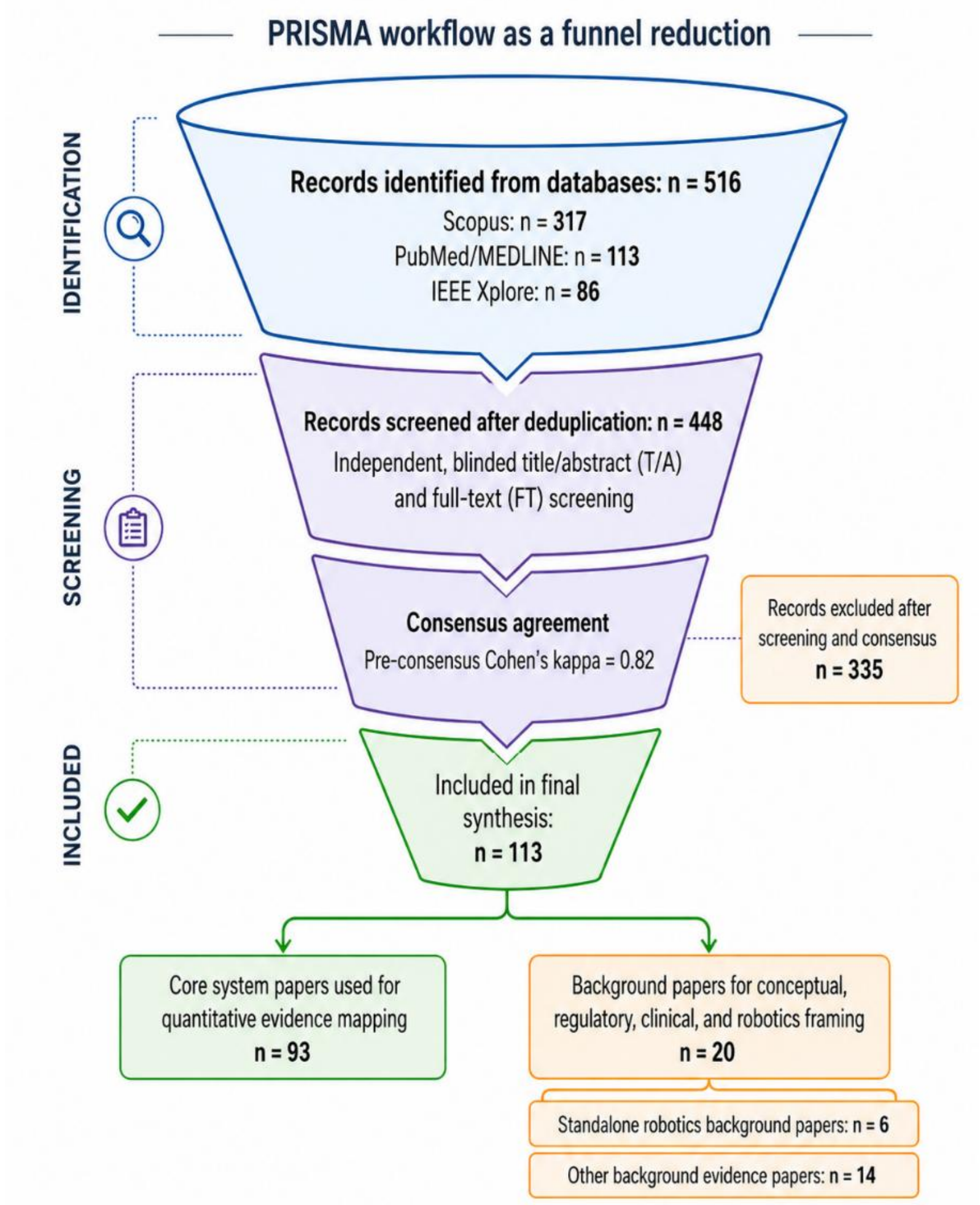

Figure 1 PRISMA-ScR-aligned flow diagram showing database identification, deduplication, screening, consensus resolution, and allocation of included records into core and background evidence streams. Abbreviation: PRISMA-ScR, Preferred Reporting Items for Systematic Reviews and Meta-Analyses Extension for Scoping Reviews.

## 2.2. Coding Framework and Terminology

Included records were charted using a prespecified coding framework designed to align data extraction with the three evidence-mapping questions. The framework combined the SIDAG architectural layers with two study-specific scales developed a priori for this review. Decision–action coupling was coded at four levels (numbered A0–A3) and validation maturity was coded at four levels (numbered M0–M3). These dimensions were defined before aggregate synthesis and were applied during independent full-text screening and preliminary extraction; disagreements or uncertain classifications were subsequently merged and resolved during consensus under senior reviewer supervision.

SIDAG formalizes and organizes functional elements recurrent across intelligent agent architectures, agentic healthcare AI, robotic sense–think–act models, RPM systems, shared autonomy, and healthcare AI governance into a healthcare-specific evidence-charting framework (10–13). It traces how systems acquire information about the patient and care context, interpret its clinical or operational relevance, select a response, and execute or trigger interventions. In high-risk healthcare and assistive robotics, this sequence requires governance mechanisms that constrain how data are handled, decisions are made, and care-relevant actions are performed (14). Data protection mechanisms alone were not considered evidence that system decisions or actions were governed. This perspective shifts the question from whether a system contains an AI agent to whether it participates in a supervised sensing-to-action pathway and, where implemented, closes the loop through outcome monitoring and feedback. For healthcare deployment, agency becomes meaningful only when sensing, inference, decision, action, outcome monitoring, and governance are linked in a way that makes system behavior traceable, interruptible, and accountable. Governance domains are reported in **Supplemental Table 15**, and paper-level governance coding is provided in **Supplemental Table 19**.

SIDAG therefore allowed us to map whether each system remained monitoring oriented or progressed toward an action-capable care architecture that linked the five SIDAG dimensions (sensing, inference, decision, actuation, and governance). These dimensions and their operational definitions are summarized in the sidebar titled The SIDAG Evidence-Charting Framework.

**[SIDEBAR] - THE SIDAG EVIDENCE-CHARTING FRAMEWORK**

SIDAG was used to code where each system stopped along the sensing-to-action chain:

- Sensing: acquisition of patient, user, robot, or environmental state
- Inference: analysis, prediction, classification, or interpretation
- Decision: recommendation, prioritization, planning, or authority allocation
- Actuation: alerting, escalation, device action, robotic movement, prompting, or intervention
- Governance: oversight, override, auditability, safety constraints, privacy, accountability, or regulatory alignment

Detailed SIDAG coding questions and examples are provided in **Supplemental Table 12** and Section 7.1 of the **Supplemental Material**.

Because the term “agency” is used inconsistently across AI, robotics, and healthcare literatures, we treated it as an operational property rather than as a broad philosophical or marketing term. Following the distinction between AI agents and agentic AI proposed by Sapkota et al. (8), we distinguish an AI agent, understood as a software or robotic actor capable of perceiving, reasoning, planning, using tools, or executing actions, from agentic AI, understood as a broader system-level capacity for goal-directed, multistep decision-making and action under constraints. Accordingly, “AI agents” refers to individual software or robotic actors; “agentic AI” refers to system-level, goal-directed decision and action; and “agentic robotics” refers to the embodied participation of robotic actors within supervised, distributed, and governed care architectures. In this review, however, agency was operationalized more narrowly as the demonstrated authority of a system to select, prepare, coordinate, trigger, or execute care-relevant action. This operational definition prevented systems from being classified as agentic merely because they used machine learning, conversational AI, or robotic embodiment. Agency had to be visible in the architecture through the coupling of decision and actuation and was assessed using the A0–A3 scale.

The A0–A3 scale captured the highest demonstrated level of decision–action coupling in each system, ranging from passive sensing, prediction, or data relay to bounded, system-governed care action. The M0–M3 scale captured validation maturity, ranging from conceptual architecture to sustained real-world deployment. Operational definitions of both scales are provided in the sidebar titled Decision–Action Coupling and Validation Maturity.

**[SIDEBAR] - DECISION–ACTION COUPLING AND VALIDATION MATURITY**

The decision–action scale used in this review is as follows:

- A0: Passive sensing, prediction, or data relay is used without decision support.
- A1: The system analyzes, alerts, or recommends; humans retain care action authority.
- A2: The system proposes, prepares, coordinates, or triggers an action that requires human authorization or supervision.
- A3: The system executes care-relevant action without immediate human confirmation, within bounded system rules.

The validation maturity scale used in this review is as follows:

- M0: The system consists only of a conceptual architecture or design proposal.
- M1: The system is a prototype or has been used in a laboratory test, simulation, or technical validation.
- M2: The system has been part of a pilot study, real-user evaluation, or deployment in a naturalistic care setting.
- M3: The system has seen sustained real-world deployment or routine clinical/service use.

The A and M levels were paired a priori because action capability and deployment maturity are analytically distinct: A system may demonstrate high decision–action coupling in a laboratory while remaining far from deployable in real care settings. Associations between embodiment and higher decision–action coupling were explored using a two-sided Fisher's exact test, comparing A0–A1 with A2–A3 systems. Odds ratios and 95% confidence intervals were reported.

Together, SIDAG, A levels, and M levels allowed us to apply the same analytical question across dashboards, AAL platforms, telecare systems, IoRT configurations, and embodied robots: Where does the system stop along the sensing-to-action chain, and what happens when inferred patient state is translated into care-relevant action? Robotic embodiment was therefore treated not as a topic separate from RPM but rather as a particularly demanding case of agentic remote care, because physical or social action requires architectural integration, human oversight, safety constraints, auditability, and accountability.

Bibliographic and conceptual information was extracted from all 113 included records using a 61-field extraction form covering bibliographic metadata, screening decisions, study characteristics, technology details, governance, limitations, gaps, and provenance. System-level SIDAG, A-level, M-level, embodiment, architecture, and governance coding was applied to the 93 core system papers used for corpus-level synthesis. AI-assisted extraction was introduced only after human consensus had finalized eligibility decisions and preliminary classifications and was limited to the postconsensus enrichment of empty or incomplete extraction fields. Human-entered values were preserved as authoritative, and all AI-generated fields were flagged for human verification before analysis.

## 3. EVIDENCE MAP: EMBODIMENT IN THE TRANSITION FROM MONITORING TO CARE ACTION

### 3.1. The Remote Patient Monitoring Baseline: Monitoring Matures Before Agency

The broader RPM/AAL corpus establishes the baseline against which the robotic signal becomes meaningful. Across 93 core systems, decision–action coupling was concentrated at A1. 67 papers (72.0%) described systems that analyze, alert, or recommend while humans retain authority over care action. Only 22 systems (23.7%) reached A2 or A3, where the system proposes, coordinates, prepares, triggers, or executes care-relevant action. Validation maturity shows a parallel concentration before deployment. 52 systems (55.9%) remained at prototype or laboratory level (M1), 30 (32.3%) reached pilot or real-user validation (M2), and only 4 (4.3%) reached sustained real-world deployment (M3).

The decision–action coupling/maturity matrix captures the central asymmetry of the field (**Figure 2**). The A1×M1 and A1×M2 cells dominate, accounting for 36 systems (38.7%) and 23 systems (24.7%), respectively. The field has therefore produced many AI-assisted prototypes and pilots but few validated agentic systems. All four M3 systems are classificated as A1 (15–18). Mature deployment is currently observed only when the technical system extends monitoring or decision support without assuming responsibility for care action. Conversely, all A2 and A3 systems remained at M2 or below. The advanced validation frontier, defined a priori as A ≥ 2 and M ≥ 2, contains only 7 systems (7.5%): 5 A2×M2 systems (19–23) and 2 A3×M2 systems (24, 25). No systems reached A2×M3 or A3×M3.

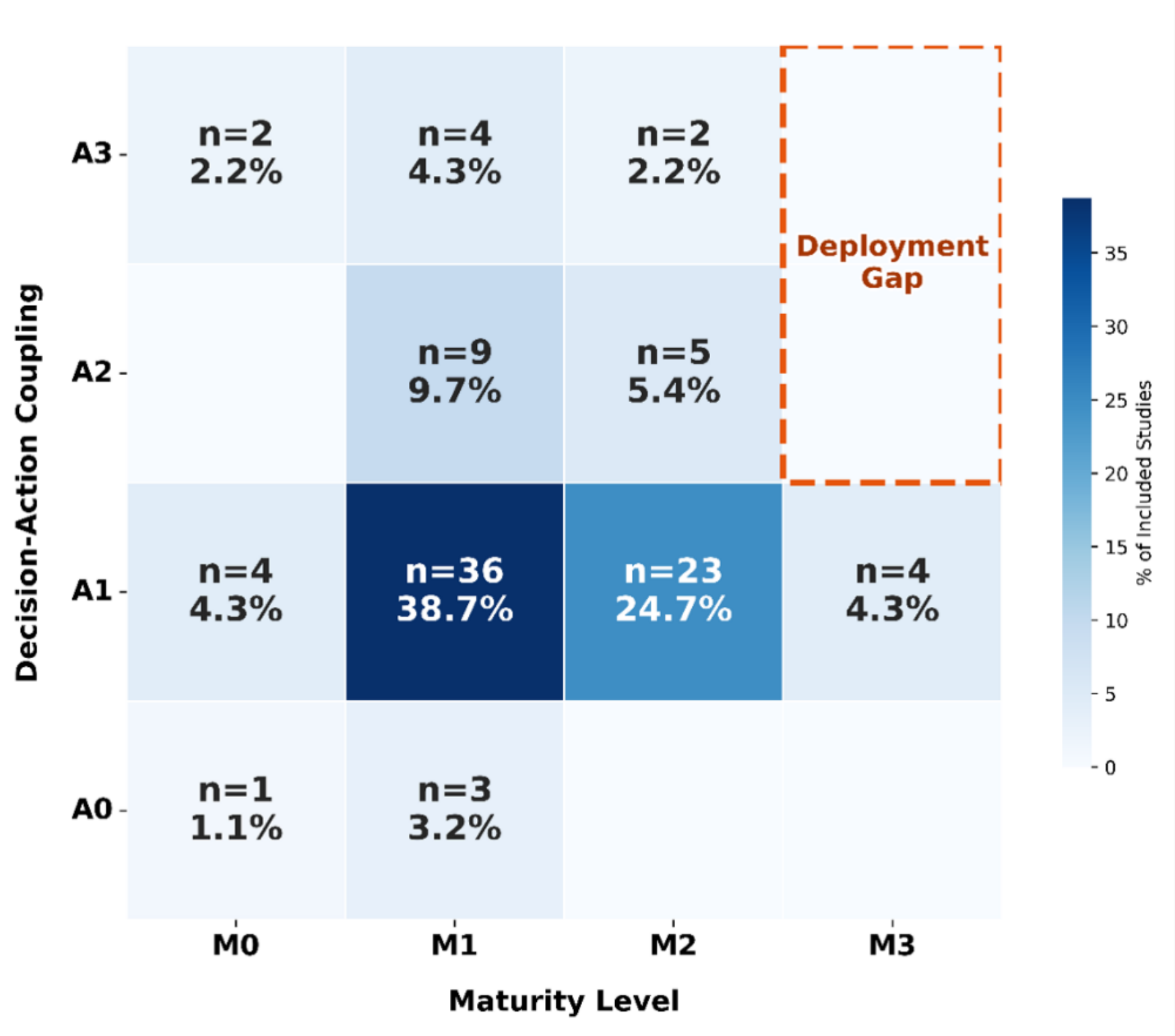

Figure 2 Autonomy maturity distribution across the 93 core systems. Cells report the number and percentage of systems in each A-level/M-level combination. The highlighted deployment gap region indicates the absence of A2/A3 systems reaching sustained deployment maturity (M3).

The lack of A2×M3 and A3×M3 systems should be interpreted carefully. The corpus does not prove a statistically significant dependency between autonomy and maturity; the pattern is descriptive and conceptual, not causal. Nevertheless, it is structurally important. One plausible interpretation is that A1 systems can mature more readily because they preserve humans as the final decision and action channel. A2 and A3 systems must also meet the broader governance requirements associated with action-capable care and integrate them into clinical workflows. Once the system can propose or execute action, deployment becomes as much a governance challenge as a technical one.

### 3.2. Embodiment as a Low-Prevalence, High-Signal Marker of Agency

Robotic embodiment is uncommon in the corpus, but it is disproportionately associated with agency. Physically embodied systems represent only 16 of the 93 papers (17.2%), reflecting the corpus's broader emphasis on digital and sensor-based monitoring. Their analytical importance derives from their association with higher decision–action. As shown in **Table 2**, nonembodied systems are overwhelmingly concentrated at A1 (61/77, 79.2%), whereas embodied systems are more evenly distributed across A1–A3. More than half of embodied systems reach A2 or A3 (9/16, 56.3%), compared with only 13 of the 77 nonembodied systems (16.9%). The relationship between embodiment and higher decision–action coupling was the clearest quantitative association observed in the corpus. Embodied systems were more frequently classified at A2 or A3 than nonembodied systems (Fisher's exact odds ratio = 6.33, $p = 0.002$).

Table 2 Embodied versus nonembodied systems

| Metric | Embodied systems ($n$ = 16) | Nonembodied systems ($n$ = 77) |
|---|---|---|
| A0 | 1 (6.3%) | 3 (3.9%) |
| A1 | 6 (37.5%) | 61 (79.2%) |
| A2 | 6 (37.5%) | 8 (10.4%) |
| A3 | 3 (18.8%) | 5 (6.5%) |
| A ≥ 2 total | 9 (56.3%) | 13 (16.9%) |
| M3 | 0 (0.0%) | 4 (5.2%) |

The observed association does not make physical embodiment a prerequisite for higher decision–action coupling, nor does embodiment alone make a robotic system capable of deciding and acting. The association shows that when inference, decision, and action acquire physical expression, the meaning and risks of agency change substantially. A nonembodied RPM system may misclassify patient state, trigger a false alert, or miss deterioration. An embodied system may move in a home, approach a person, mediate a care interaction, prompt medication or rehabilitation, support mobility, manipulate objects, or trigger a physical or social response. Errors can therefore extend beyond informational harm and produce physical, relational, and clinical consequences.

Embodiment agency therefore requires carefully orchestrated architectures that couple perception and reasoning with physical and social capabilities, because each function affects the operation and safety of the others. Consistent with robotics background literature, physical actuation must remain subject to safety-critical constraints throughout this integrated process to support situated intelligence and action (26). Wearable and soft robotic assistance further illustrates why embodied agency must adapt in real-time to patient-specific biomechanics while maintaining comfort and safe force transmission (27). For this reason, embodiment marks the point at which care action acquires materially and socially consequential.

The heterogeneity of the robot subset is analytically useful, because it reveals distinct pathways from monitoring to embodied action. Social and dialogue-oriented robotic systems translate inference into interaction or prompting (19, 28, 29)., while telepresence systems connect monitoring with navigation and remote presence (20, 30, 31). In IoRT and multirobot configuration, coordination becomes part of the care action (11, 32, 33), whereas humanoid and rehabilitation-oriented systems combine monitoring, interaction, and embodied support (23, 34, 35). **Figure 3** provides a visual anchor for these pathways, spanning social robotic interfaces, assisted telepresence, and modular healthcare robotics platforms.

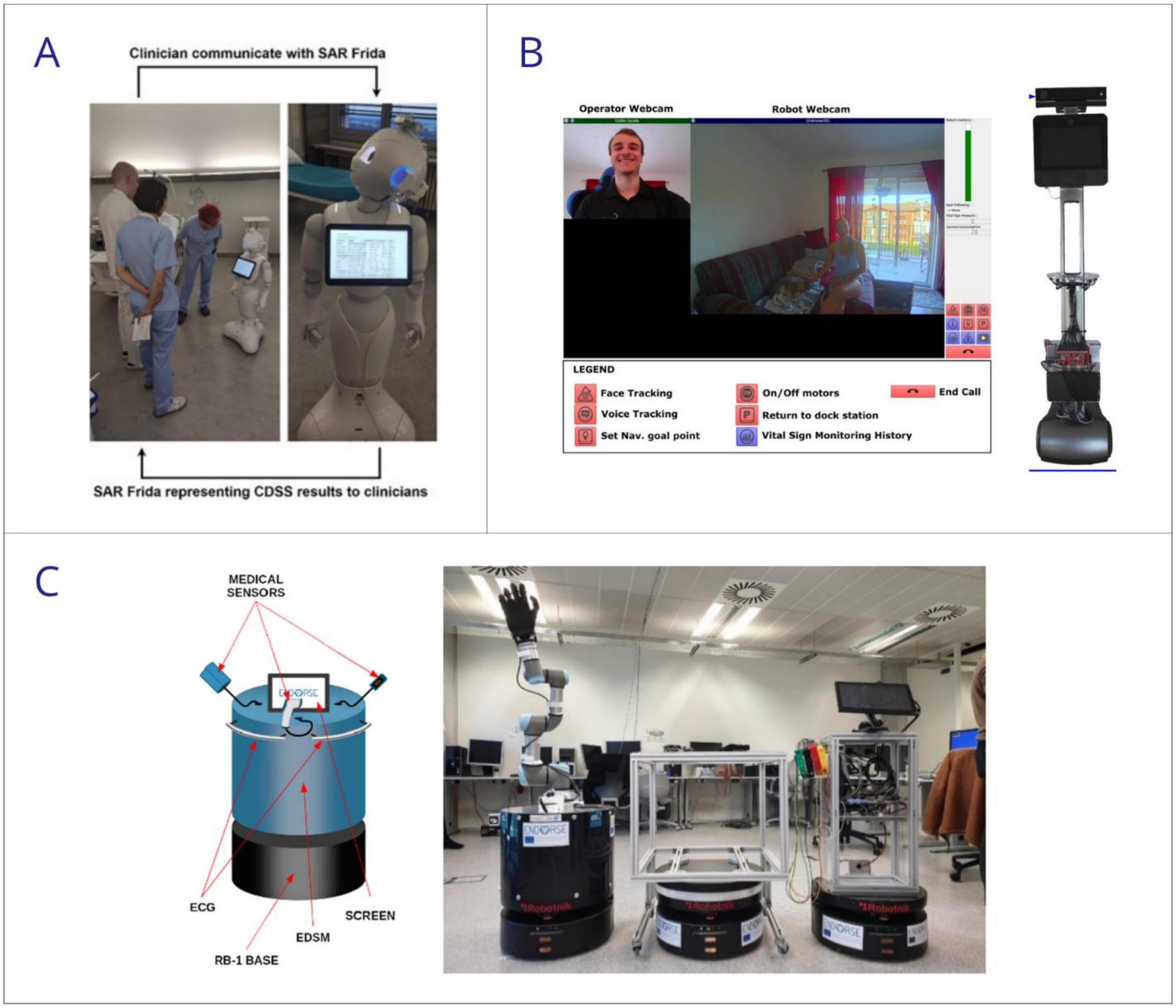

Figure 3 Representative embodied systems from the reviewed corpus. (*a*) The Frida SAR used as an embodied interface for clinical decision support and patient-monitoring data. Panel adapted from Reference 19 (CC BY 4.0). (*b*) The SAM augmented telepresence robot for assisted autonomous navigation and remote home care interaction. Panel adapted from Reference 20 (CC BY 4.0). (*c*) ENDORSE modular healthcare robotic platforms discussed in the HERON work (35). Panel adapted from Reference 35 (CC BY 4.0). Abbreviations: CDSS, clinical decision support system; ECG, electrocardiogram; EDSM, [Electronic Diagnostic Support Module]; HERON, Healthcare Robotics' Ontology; SAR, socially assistive robot.

### 3.3. The Robotic Deployment Gap: From Stand-Alone Capability to Governed Care Architecture

The same evidence that makes robots analytically important also exposes the deployment gap. No embodied system reaches M3 sustained deployment. The small number of M3 systems limits statistical inference, but this pattern remains central to the review's argument. Embodied systems are associated with higher decision-action coupling without a corresponding increase in deployment maturity. Mature deployment is currently confined to conservative A1 tools, while action-capable systems remain prototypes or pilots. This ceiling concerns sustained deployment within the care architectures captured by the review and does not preclude embodied social robots from reaching users through consumer or other nonregulated care pathways.

ElliQ (36) provides an instructive boundary case. It has reached commercial and public sector deployment as a proactive companion robot for older adults, exhibiting social-assistive agency through self-initiated interaction and the personalization of reminders, wellness support, and communication based on users' routines and preferences. ElliQ nevertheless remains positioned as a consumer device and it is not intended to detect, diagnose, treat, triage, or manage clinical deterioration. Even with this bounded action-space, deployment requires careful calibration and support. Users may need guidance about the robot's role and technical assistance, and some may perceive proactive interactions as intrusive. The study also describes an optional weekly report that users may share with clinicians, while clinical interpretation and any resulting action remain external to the system (36). Social-assistive agency can therefore mature commercially when its action space focuses on companionship and wellness support and clinical authority stays external.

Viewed against this boundary case, the deployment gap appears to concern the transition from local robotic capability to system-level readiness. Adjacent AAL and robotics-control studies identify complementary mechanisms for supporting this transition through distributed service coordination (37), simulation-based validation in inhabited environments (38), and formal verification of safety and coordination in multirobot settings (39).

As shown in **Table 3**, only seven systems combine $A \geq 2$ with $M \geq 2$. The embodied subset comprises a socially assistive robot (19), a telepresence robot (20), and a humanoid/IoT platform (23). Across these systems, actions remain bounded and the care authority is shared through human supervision or explicit governance. Within the reviewed corpus, governed and ecosystem-embedded action therefore represents the current empirical frontier of embodied agentic RPM.

Table 3 Seven near-deployment systems

| Reference | Title | A level | M level | Care function | Robot | Governance |
|---|---|---|---|---|---|---|
| 19 | Feasibility of a computerized clinical decision support system delivered via a socially assistive robot during grand rounds: a pilot study | A2 | M2 | Social assistance | Social | Multiple |
| 24 | AI-driven patient monitoring and mental health analytics platform | A3 | M2 | Monitoring | None | Multiple |
| 20 | Toward enhancing the autonomy of a telepresence mobile robot for remote home care assistance | A2 | M2 | Social assistance | Telepresence | Multiple |
| 25 | DOMUS: a domestic ontology managed ubiquitous system | A3 | M2 | Rehabilitation support | None | Multiple |
| 23 | Smart Care: an IoT platform with humanoid robots for diabetes management in children | A2 | M2 | Rehabilitation support | Humanoid | Human-in-the-loop |
| 21 | Personalized multimodal AI and AR System for remote cognitive assistance in aging-in-place | A2 | M2 | Social assistance | None | Multiple |
| 22 | My-AHA: software platform to promote active and healthy ageing | A2 | M2 | Rehabilitation support | None | Multiple |

The evidence map identifies the structural tension of the field: Deployment maturity accumulates around conservative monitoring systems, whereas embodied and agentic systems concentrate action capability but remain below sustained deployment. The next section examines how this tension emerged over time by tracing the evolution from cloud-based, rule-driven monitoring to edge–hybrid inference and more recent forms of agentic and embodied care systems facing unresolved governance requirements.

## 4. FROM PREDICTIVE MONITORING TO EMBODIED ACTION

RPM and AAL systems have not evolved along a single axis. The corpus shows neither a simple replacement of cloud dashboards by robots nor a linear progression from rules to machine learning and autonomy. The clearest temporal change concerns the location and distribution of computation. Hybrid platforms increased from 9 of 27 systems (33.3%) in 2018–2020 to 10 of 22 (45.5%) in 2021–2023 and 22 of 44 (50.0%) in 2024–2026, while dedicated edge platforms appeared after 2020. Decision–action coupling did not follow the same trajectory: Systems classified at A2 or A3 represented 7 of 27 (25.9%), 3 of 22 (13.6%), and 12 of 44 (27.3%) across the three intervals, respectively. The corpus therefore indicates changing architectural emphases rather than successive stages toward autonomy.

Embodiment is distributed across all three intervals and is treated as the cross-cutting analytical dimension defined in Section 2 and examined empirically in Section 3. Its relevance here lies in how different architectural configurations accommodate embodied action and the associated requirements for supervision and governance.

**Figure 4** summarizes these changes in architectural emphasis, from monitoring-centered configurations with early distributed and embodied exceptions, through the expansion of edge–hybrid processing, to the consolidation of heterogeneous distributed ecosystems supporting bounded forms of care-relevant action. Paper-level architecture, decision–action coupling, maturity, embodiment, and governance classifications are included in **Supplemental Table 19**, while the physically embodied subset is reported separately in **Supplemental Table 21**. The 2024–2026 interval includes records indexed through April 2026 and therefore does not represent a complete three-year publication window.

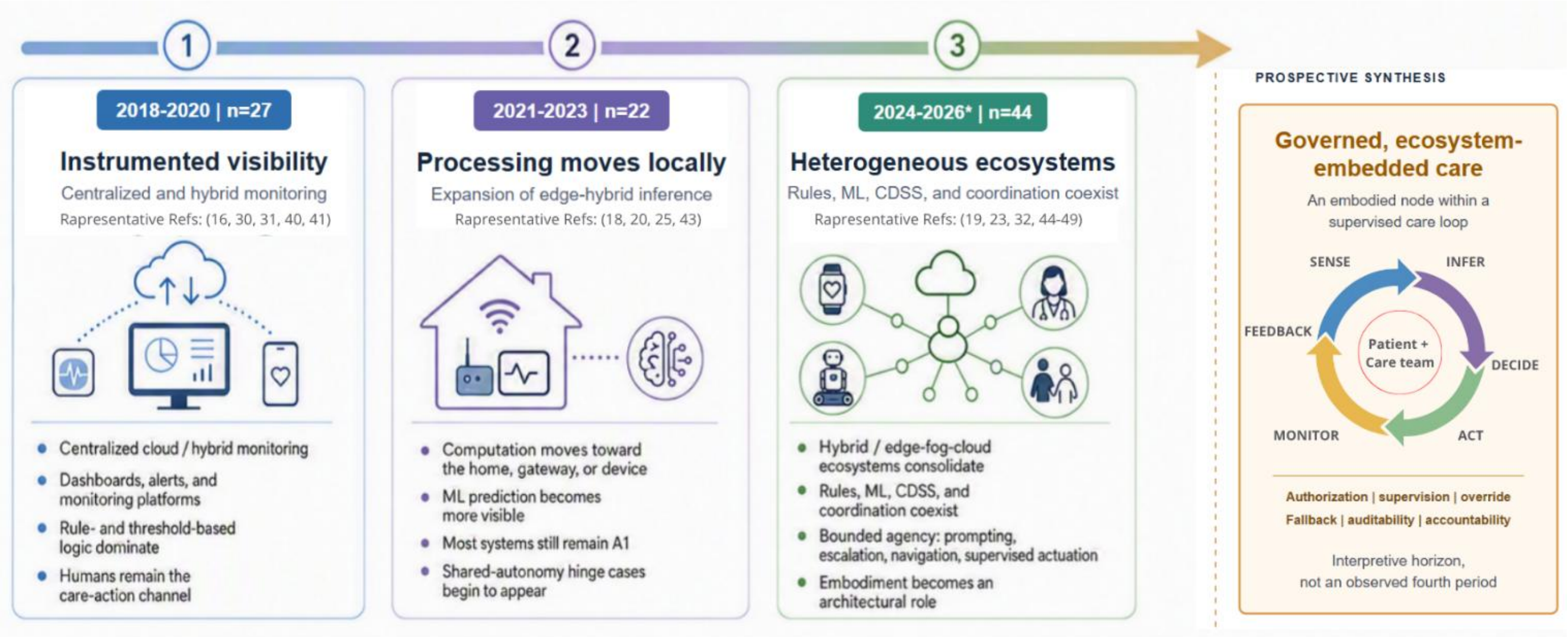

Figure 4 From predictive monitoring to governed embodied action: the architectural evolution of AI-enabled RPM, AAL, and assistive robotics between 2018 and 2026, from cloud-based visibility to edge–hybrid inference and distributed, decision-capable care ecosystems. The clearest temporal signal is more distributed computing rather than a monotonic rise in decision–action coupling or embodiment. Abbreviations: CDSS, clinical decision support system; ML, machine learning.

## 4.1. Instrumented Visibility and Centralized Monitoring

The first period, 2018–2020, is best understood as the era of instrumented visibility. The dominant technological problem was how to acquire the patient, user, and environmental state and make it available to clinicians or caregivers. Cloud and hybrid platforms were equally common (9/27 each, 33.3%), while dedicated edge processing was absent. Across these configurations wearable, ambient, or device-level signals were typically routed to cloud services, dashboards, or local monitoring platforms for humans' interpretation (40, 41). The corresponding AI profile was conservative. Rule-based and threshold mechanisms accounted for 17 of the 27 systems (63.0%), and 19 of the 27 systems (70.4%) were classified as A1. Their contribution lay primarily in improving detection and clinical visibility while care authority remained with humans.

Wearable AI monitoring in assisted-living communities (16) illustrates this A1 pattern within the period. The same human-led configuration subsequently reached sustained deployment in oncology symptom monitoring (15), chronic heart failure telemonitoring (17), and electronic health record (EHR)–integrated home spirometry after lung transplant (18), indicating the persistence and maturity of this architectural model beyond the first interval. Embodiment also appeared early, although not as mature care infrastructure. Robot-related systems such as the elderly AAL robot sensor architecture (30) and wearable-triggered tele controlled assistance (31) already connected robotic action to sensing, servers, planning, or teleoperation. These systems revealed an architectural tension already present at the beginning of the review period: Physical action could be introduced before the surrounding care ecosystem was ready to govern it. The robot could move, mediate, or respond, while the broader RPM/AAL pathway remained prototype level, partially validated, or weakly connected to clinical workflows and accountability.

The same distinction is important for interpreting locally bounded robotic systems. An assistive device may combine sensing, control, and actuation while remaining only weakly integrated with a broader monitored care pathway. The cane-type fall prevention robot (42) provides a later example of this boundary: It has clinically meaningful physical embodiment, including motion sensing, vibrotactile feedback, lighting response, and onboard actuation, but its contribution remains local and technical. Such systems should show what embodied capability looks like before ecosystem integration and why embodiment alone is not equivalent to agentic care.

### 4.2. Edge–Hybrid Inference and Local Processing

Between 2021 and 2023, computation increasingly moved closer to the point of care. Hybrid edge–fog–cloud architectures rose to 10 of 22 systems (45.5%), and dedicated edge platforms appeared (3/22, 13.6%). Papers increasingly addressed the limitations of centralized cloud monitoring for timely, bandwidth-efficient, and privacy-sensitive processing. Inference consequently moved toward the home, gateway, device, or robot, as illustrated by edge–cloud clinical prediction and autonomous telepresence systems developed during this period (20, 43).

The AI profile changed accordingly. Machine learning prediction rose to 8 of 22 systems (36.4%), becoming more prominent than in the first period. This shift is important but should not be overread. Predictive machine learning increased inferential sophistication and the systems could model deterioration, classify activity, identify anomalies, or detect patient-specific risk more flexibly than threshold logic alone. Yet the A-level distribution remains conservative: 19 of 22 systems (86.4%) were classified as A1, while only 3 of 22 (13.6%) reached A2 or A3. Systems increasingly incorporated predictive methods without a corresponding increase in the authority delegated to them; most outputs remained directed to humans for interpretation and response. Better models did not automatically close the care loop. A patient-criticality-aware edge–cloud architecture could perform clinical prediction closer to the data source while still routing its output to human decision-makers (43). Similarly, EHR-integrated home spirometry combined automated prompting and data collection with threshold-based alerts displayed on clinical dashboards, leaving care authority with clinicians (18). A predictive RPM platform may therefore be technically advanced but architecturally conservative if its outputs stop at a dashboard or alert. Conversely, a simpler rule-based or ontology-driven system may be more agentic if it coordinates an action pathway under defined supervision, as illustrated by the bounded domestic actuation implemented in DOMUS (domestic ontology managed ubiquitous system) (25).

The question for this review is therefore where AI output goes and what authority it carries. Does it remain informational, or does it become coupled to decision, actuation, escalation, or embodied support?

The telepresence robot (20) is the hinge case for this by translating local computation into supervised embodied action. The user or operator defines the goal, and the robot executes bounded movement while retaining fallback control pathways. Its validation in a real home demonstrates a shared autonomy configuration with an explicit allocation of human and robotic authority. Adaptivity-as-a-service (37) extends this principle by modelling assistance through services coordinated across a distributed AAL architecture. Formal and stochastic modeling of multirobot service settings (39) similarly treats human behavior and coordination as part of the system. Together, these studies locate intelligence at the level of the distributed care loop, where local computation and robotic action remain coordinated with human supervision and fallback.

### 4.3. Consolidation of Hybrid Ecosystems and Heterogeneous Decision Logic

During the third period, 2024–2026, hybrid configurations accounted for 22 of 44 systems (50.0%), while dedicated edge platforms accounted for 5 of 44 (11.4%). Decision logic remained heterogenous. Rules and thresholds were used in 24 of 44 systems, (54.5%), machine learning and adaptive methods in 13 of 44 systems (29.5%), and smaller groups incorporated clinical decision support systems (CDSS) or autonomous and multiagent logic. This coexistence indicates continued methodological plurality across the field. Systems combine rule-based and statistical methods with clinical decision support, planning, or coordination according to the function of their outputs and where action is expected to occur.

Recent systems illustrate this plural AI ecology: TeleXAI-R combines transformer-based risk stratification with explainability and confidence-aware alerting (44), while edge–AI enables local anomaly detection with clinician-facing dashboard alerts (45); fog–edge Internet of Medical Things (IoMT) architectures optimize latency and threat resilience (46). Other systems connect decision logic more directly to clinical workflows. Chronic heart failure platforms integrate CDSS, Fast Healthcare Interoperability Resources (FHIR) workflows, alerts, and care plans (47); conversational AI conducts heart failure follow-up while nurses retain treatment authority (48); and local large language model (LLM) and multiagent elder care systems add voice, emotion recognition, reminders, and caregiver alerts without assuming care authority (49).

The observed diversity can be interpretated in light of the different computational and control requirements of risk estimation. A monitoring system can tolerate a probabilistic prediction if a human interprets the result. An embodied or semiautonomous system must also decide when to interrupt, prompt, move, escalate, hand over, or remain silent. Coupling AI outputs to care-relevant action shifts the design problem from prediction to authority allocation and control under uncertainty, making decision logic inseparable from the architecture question.

The robotic system from this period should be read as architectural roles associated with specific care functions. The socially assistive robot used to deliver FHIR-integrated clinical decision support during hospital rounds (19) serves as a clinical interface connecting patient-specific information with clinician workflow. The Smart Care diabetes platform (23) uses a humanoid robot as a home feedback node that collects biometric and self-reported information, links it to a disease-management hub, and returns personalized feedback to the patient, supporting day-to-day diabetes self-management . The human–robot–robot assistance system (32) extends this configuration toward IoRT orchestration. In this system, bed and privacy-preserving thermal sensing detect the patient's intention to exit the bed, while a mobile robot interacts with the patient and manipulator-based assistance enables object handover, aiming to reduce unsafe bed-exit events and support activities of daily living. Together, these systems address the shared architectural and clinical challenge of moving intelligence beyond the dashboard and into the care environment while maintaining human supervision.

Agentic AI becomes relevant when heterogeneous decision logic is coupled with care-relevant action. Open-ended autonomous agents have not reached broad deployment within the corpus, and the observed forms of agency remain bounded to informational support, care coordination, and supervised embodied action As established in Sections 2 and 3 and illustrated in Section 4.2, the degree of agency depends on where system outputs are directed and what authority they carry. Embodiment makes this coupling particularly consequential because digital outputs acquire physical or social expression within the patient's environment.

Earlier and adjacent robotics research identifies additional capabilities relevant to this transition. Digital-twin validation of Pepper in a simulated smart-home environment (38) shows how integrated robotic subsystems can be evaluated before physical deployment, while accounting for computational, environmental and interactional failures. Advanced soft robotics work (50) highlights the need for adaptive actuation and human–robot interaction mechanisms when assistance has physical consequences. Artificial theory-of-mind trajectory prediction (51) shows that implicit human intent must be inferred when people and robots share space without explicit verbal commands, while robotic walking assistance illustrates real-time adaptation to kinematic and physiological constraints (52). The Healthcare Robotics' Ontology (HERON) (35) provides the semantic layer needed for communication, collaboration, and safety across healthcare robotics. Together, these examples characterize embodied care intelligence as multimodal, anticipatory, interactive, and safety constrained.

### 4.4. Clinical Use Cases: From Remote Monitoring to Situated Support

The clinical implications of these architectural shifts become clearer in home care use cases. Conventional RPM, primarily supports earlier recognition of heart failure instability, chronic disease instability, abnormal activity, and other signs of clinical deterioration (15, 17, 18, 53). In embodied RPM/AAL, the expected benefit shifts from earlier detection to situated support. The system is no longer only asking whether something is wrong; it begins to support what should happen next. **Figure 5** provides a visual overview of this shift, showing how embodied RPM/AAL extends clinical visibility toward situated support across key home care use cases.

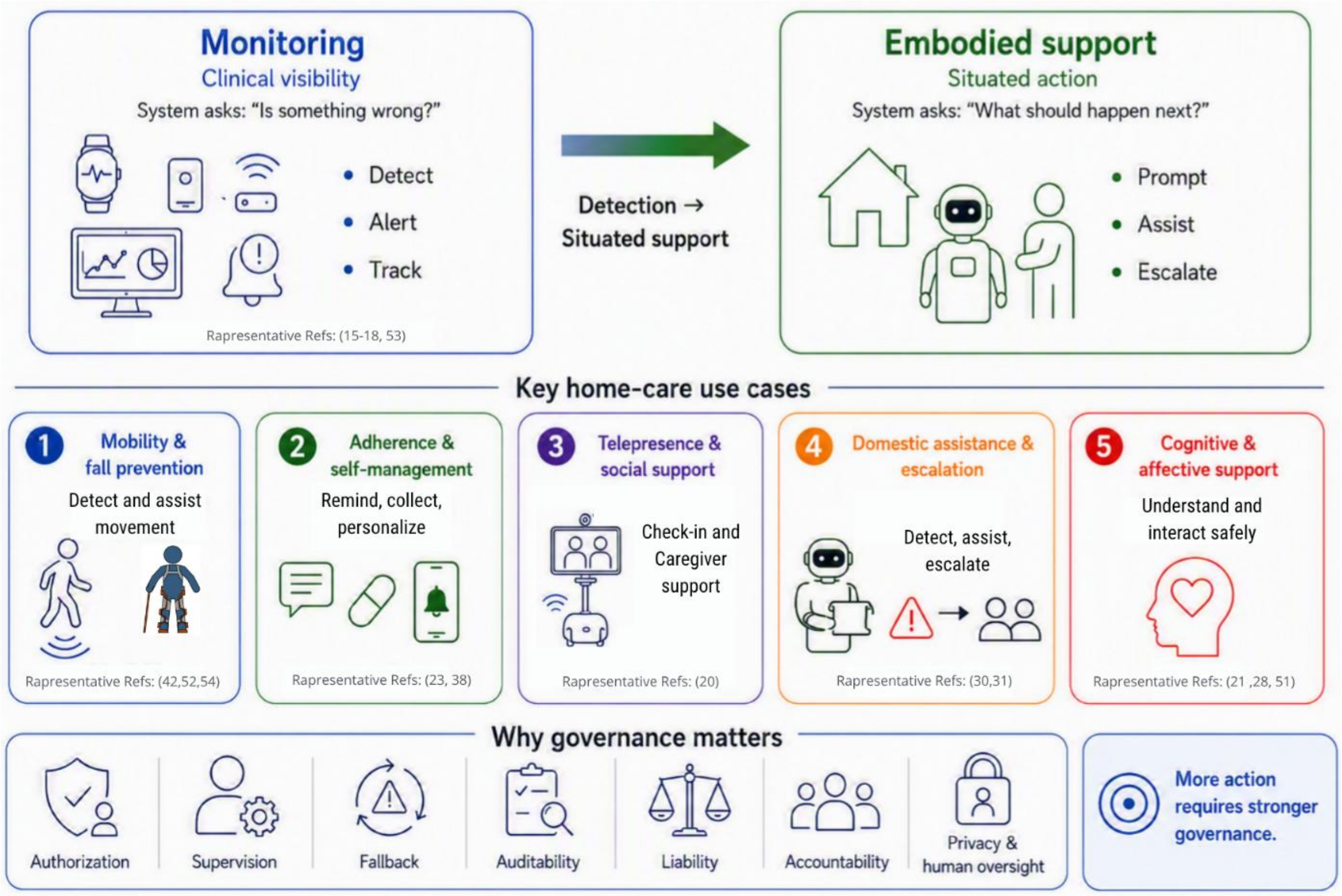

Figure 5 Overview of how embodied RPM/AAL extends clinical visibility toward supervised assistance in key home care use cases, increasing the need for explicit governance. Embodiment shifts care from earlier detection to meaningful action in daily life. Representative references include studies from the core corpus and adjacent robotics literature. Abbreviations: AAL, ambient assisted living; RPM, remote patient monitoring.

Fall prevention and mobility support illustrate this shift clearly. Wearable and ambient system can detect instability, inactivity, or a fall and notify caregiver (54). Embodied system can also provide local assistance. A cane-type robot can provide vibrotactile feedback, lighting support, or motion stabilization (42). Adaptive assistance system can modulate robotic support in response to fatigue, instability, or movement state estimates (29), while walking-assistance work in the robotics background literature shows that safe physical cooperation requires real-time adaptation to the user's gait and kinematic limits (52). These capabilities support the clinical objectives of reducing fall risk, preserving mobility, and sustaining activities of daily living.

Medication adherence and self-management provide another pathway toward situated support. Nonembodied RPM platforms can remind, alert, or display adherence data. Social and humanoid robots make this loop more interactive by asking questions, collecting self-reported symptoms or behaviors, and returning feedback in a familiar home context. In the Smart Care diabetes IoT platform (23), the robot serves as the patient-facing interface connecting biometric measurement and self-reported behaviors with a disease management hub and personalized feedback. Similarly, medication-reminder dialogue systems (28) show how social embodiment can support adherence routines. Integration with clinical oversight and escalation pathways determines how these systems participate in the broader care architecture.

Telepresence and social assistance address isolation, caregiver burden, and access to remote support alongside clinical monitoring. The telepresence robot described by Laniel et al. (20) combines remote presence with bounded autonomous navigation under human command. Its mobile interface allows caregivers or clinicians to interact with and observe the home environment, supporting remote check-ins and situational awareness. This configuration depends on clearly defined privacy and control boundaries, including fallback control and human authority.

Domestic assistance and emergency coordination are the most action-oriented use cases. AAL robot sensor architectures (30) and IoRT systems (32) connect environmental sensing, intent recognition, dialogue, mobile robot action, and sometimes manipulation. They can detect unsafe attempts to leave the bed and ask whether help is needed. Robotic components may then provide object assistance or notify staff or relatives. These examples show how embodiment changes the risk profile, intervening in a vulnerable moment in the patient's environment.

Finally, adaptive and cognitive support illustrates how clinical value also depends on personalization and anticipation. Adaptive robotic dialogue systems can tailor medication reminder interactions to user responses and changing conversational conditions (28), while multimodal AI and augmented reality systems can provide personalized remote cognitive assistance for aging-in-place (21). Artificial theory-of-mind trajectory prediction illustrates how safe interaction may also require anticipating what a person is about to do when humans and robots share space (51). Together, these approaches aim to preserve autonomy through assistance that responds to the user's behavior and context.

Across these use cases, embodied systems extend clinical visibility into supervised support for daily functioning. As action becomes more care-relevant, the SIDAG governance safeguards defined in Section 2 become increasingly central. Section 5 examines how these requirements are implemented and reported across the reviewed systems.

## 5. GOVERNANCE READINESS: FROM CARE ACTION TO ACCOUNTABLE AUTONOMY

The clinical use cases discussed above create a natural transition from architecture to governance. Once RPM/AAL systems move from monitoring to situated support, the key question is whether care-relevant action is authorized, supervised, explainable, interruptible, and accountable. A system that displays an alert can leave responsibility with clinicians or caregivers. When system prompts medication, actuates a domestic device, escalates a crisis, or coordinates robotic assistance, it redistributes authority within the care loop. Across the 93 core systems, 24 papers (25.8%) either had no identifiable governance mechanism or provided insufficient information for governance coding. The absence or nondetermination of reported governance mechanisms should be interpreted as a governance readiness and reporting gap, not as proof that safeguards were absent in the prototypes. Nondocumentation nevertheless consequential because deployment readiness cannot be assessed when the relevant SIDAG safeguards are unspecified.

A1 systems, although technically conservative, have the largest governance gap. 20 of 67 systems (29.9%) had no identifiable governance mechanism or insufficient governance information. A2 systems showed the lowest gap, with only 1 of 14 (7.1%) lacking governance documentation. This lower proportion may reflect the need for systems coordinating executable actions to make authorization workflows visible. 2 of 8 A3 systems (25.0%) had no identifiable or determinable governance mechanisms despite system-governed action, although the small denominator limits interpretation. Governance therefore becomes relevant whenever AI output influences care decisions, escalation, workload, or patient behavior. **Figure 6** summarizes how the identified governance features are distributed across autonomy levels and between embodied and nonembodied systems.

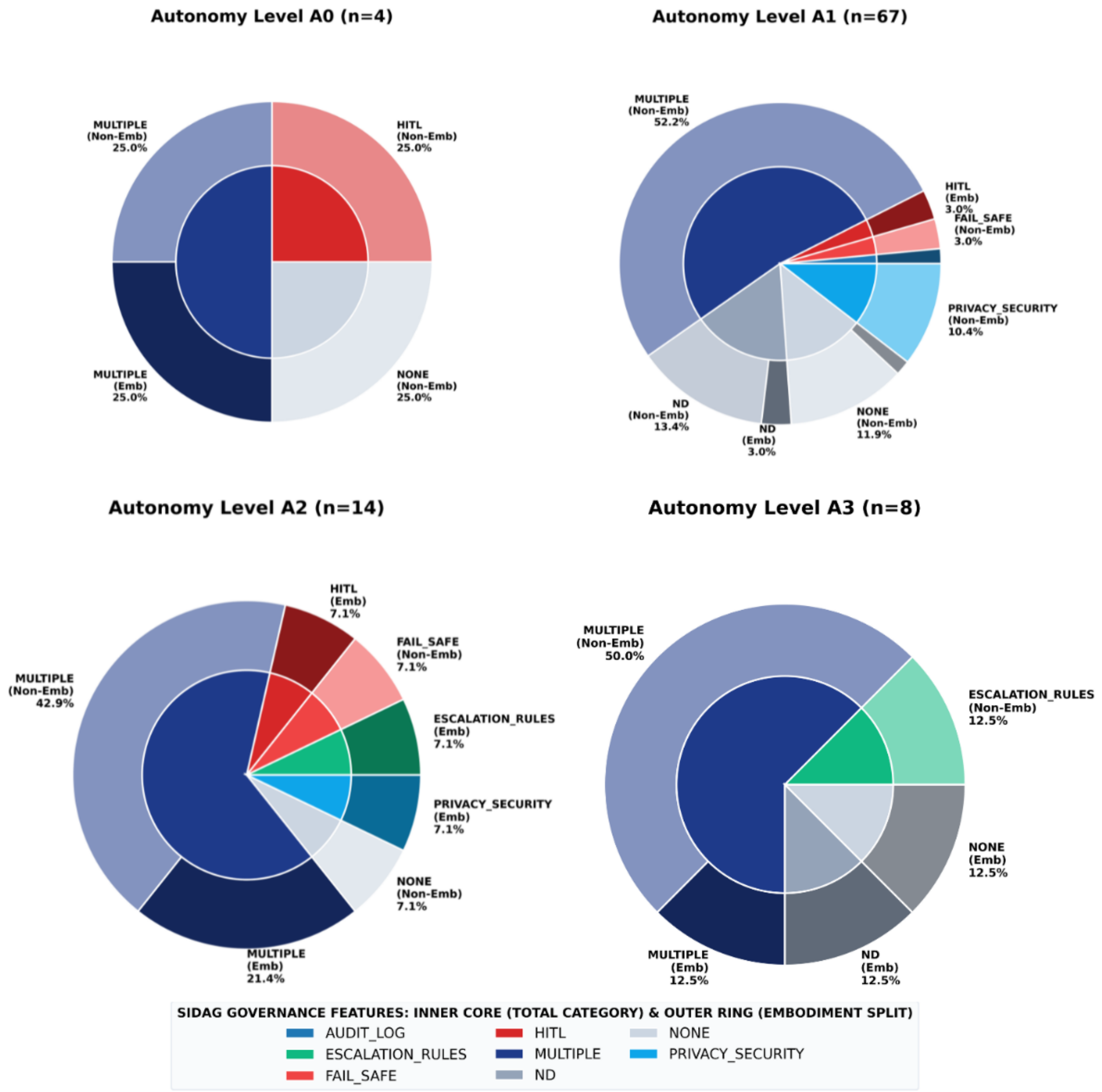

Figure 6 Governance feature subdivisions by robotic embodiment. Inner rings show governance categories, and outer rings distinguish embodied and nonembodied systems. ND indicates that the paper provided insufficient information to determine whether a governance mechanism was present; "none" denotes that the paper had no identifiable governance mechanism (not merely that one wasn't reported or that the paper provided insufficient information). Detailed governance mechanisms and coding criteria are reported in **Supplemental Table 15**. Abbreviations: HITL, human-in-the-loop; ND, nondeterminable.

Among the action-capable systems with real-user validation, the pattern becomes more concrete. All 7 systems reaching both $A \geq 2$ and $M \geq 2$ document at least one governance mechanism. Six reported multiple mechanisms, and one reported human-in-the-loop governance. This pattern is consistent with the broader RPM and telemedicine literature, which identifies technical performance, workflow integration, role allocation, trust, interoperability, and accountability across the care pathway (55, 56). The observed association does not establish causality, because pilot studies may simply encourage more detailed documentation through ethics review, institutional oversight, and user testing. Even so, systems closest to deployment tend to specify how action is authorized, supervised, or constrained. The distribution of decision authority was a more informative discriminator than whether a system was cloud based, edge based, hybrid, or robotic. 5 of the 7 near-deployable systems are A2, where the system proposes or prepares action while a human remains in the authorization loop. A3 systems carry a higher governance burden because autonomous escalation and domestic actuation require clear oversight, fail-safe behavior, auditability, and override to become credible care infrastructure (24, 25).

The European Union's Artificial Intelligence Act gives this argument regulatory urgency and provides a readiness lens for the analysis, while formal legal classification remains outside the scope of the review. Depending on the intended use, product status, and deployment context, AI systems affecting health, safety, vulnerable users, or regulated products may fall under high-risk obligations. Relevant requirements correspond closely to the gaps observed here and include risk management, technical documentation, logging, information to deployers, human oversight, robustness, cybersecurity, and accuracy (57). The review does not assess legal compliance. Its findings show that many papers do not yet report the evidence needed to assess compliance readiness.

The governance implications become especially visible in embodied RPM/AAL because physical and social actions can turn weakly documented authority into a direct safety issue: Who can stop the robot, reverse an automated action, interpret an escalation, or review what happened afterward? Progress toward accountable autonomy therefore requires systems that close the loop from patient state to care action while making authority, traceability, and responsibility explicit.

These findings support agentic robotics as the prospective design horizon developed in the concluding section, with governed, ecosystem-embedded care robots whose actions are selected through clinically validated policies and remain subject to explicit human authority.

## 6. CONCLUSIONS AND FUTURE DIRECTIONS

This review was guided by three evidence-mapping questions concerning how AI-enabled RPM, AAL, telecare, and robotic care systems are architecturally configured; how embodied systems are positioned across decision–action coupling and maturity levels; and which patterns characterize systems that approach deployable care action. Taken together, the findings show a field moving from monitoring toward action but not yet toward fully autonomous care. Some systems reach bounded A3 action, but none demonstrate open-ended autonomous care in sustained real-world deployment. Current maturity remains concentrated in conservative A1 architectures, where AI supports monitoring, prediction, alerting, or decision support while humans retain responsibility for clinical action. Findings from the small embodied subset require cautious interpretation, physical embodiment is neither a prerequisite for higher decision–action coupling nor sufficient to make a system deployable. More specifically, the evidence shows that embodied systems are disproportionately associated with higher decision–action coupling. Unlike a dashboard, a robot can turn AI output into situated physical or social behavior through speech, movement, prompting, navigation, manipulation, and interaction with patients or caregivers. Embodiment therefore makes the transition from monitoring to action visible while raising the burden of safety, oversight, validation, and accountability.

The temporal trend supports this interpretation. Earlier RPM/AAL systems made care more visible through cloud monitoring, dashboards, and rule-based alerts. Edge–hybrid architectures then moved inference closer to patients and homes, while predictive machine learning improved detection without necessarily changing authority. More recent systems have explored distributed and decision-capable ecosystems in which AI models, sensors, devices, robots, clinicians, caregivers, and users coordinate bounded care action. The field is therefore redistributing intelligence across a care ecosystem rather than progressing linearly from rules to machine learning and autonomous control.

The evidence map supports the governed, ecosystem-embedded care robot as a prospective design horizon. We use the term "agentic robotics" to describe this prospective horizon beyond the maturity levels observed across the corpus. Agentic robotics refers to a multiagent care architecture in which artificial agents, robotic embodiments, clinicians, caregivers, and patients share perception, decision support, action, and accountability across a supervised care loop. Within this architecture, robots operate through clinically validated action policies appropriate to their intended use and level of risk, with explicit human authority over clinically consequential decisions. Their agency derives from participation in a context that makes action informed, constrained, traceable, and accountable. By integrating patient and environmental sensing, distributed inference, clinical and caregiver knowledge, local interaction, and governance, such architectures could strengthen perception and adaptation in unstructured home environments. In this model, embodiment provides AI with a situated interface to daily life, while ecosystem integration prevents situated action from becoming isolated, opaque, or unsafe.

This design horizon extends beyond the maturity demonstrated by the reviewed systems. Those with the strongest decision–action coupling remain below sustained deployment, while background reviews identify continuing translational barriers involving data quality, interoperability, clinical workflow, digital inclusion, cybersecurity, scalability, and multicenter validation (55, 58–60).

Progress toward agentic robotics requires governance to be designed and reported as an architectural component using the safeguards formalized in SIDAG. Explainability must also support interpretation of system output by clinicians, caregivers, and users. Validation should advance from short-term pilots to longitudinal and multisite evaluation of A2 and A3 systems, with embodied technologies assessed as a component of care ecosystems instead of isolated robot prototypes. Technical performance must ultimately be linked to clinical meaningful outcomes, including preserved mobility, improved adherence and self-management, reduced caregiver burden, stronger social connection, and timely escalation.

The strongest conclusion of this review is that embodied AI becomes clinically credible only when it is connected, supervised, and accountable. Accountable autonomy therefore defines the most promising horizon for agentic robotics. Ecosystem-embedded robotic and RPM/AAL systems should act through context aware and governed pathways, closing the loop from patient state to care action while keeping human authority, traceability, and responsibility explicit.

## DISCLOSURE STATEMENT

I.R. and F.C. are affiliated with CoAImed S.r.l. The authors declare that this affiliation did not influence study selection, coding, interpretation, or manuscript preparation.

## AUTHOR CONTRIBUTIONS

I.R. and L.G. conducted independent blinded screening and preliminary extraction of the principal classification fields. I.R. led postconsensus extraction completion, data curation, statistical analysis, and manuscript drafting. L.G. contributed to full-text eligibility assessment, classification reconciliation, extraction checking, and interpretation. F.C. supervised the project, participated in consensus resolution, validated the analytical framework and classifications, and critically revised the manuscript.

## ACKNOWLEDGMENTS

Data extraction was supported by AI-assisted prepopulation of extraction fields after human consensus decisions had finalized study eligibility. The workflow used a persistent LLM Wiki pattern inspired by Karpathy and related AI-wiki methods (https://gist.github.com/karpathy/442a6bf555914893e9891c11519de94f).
All AI-generated extraction fields were explicitly flagged and independently verified against the original full-text articles by the primary reviewer before inclusion in the final analysis.

OpenAI's ChatGPT (GPT-5.5) was used to assist with English language editing, improve readability, and refine the wording and organization of selected sections of the article. ChatGPT was also used to refine the visual style of selected conceptual figures. The conceptual design of these figures was initially developed using Google's NotebookLM. The textual content and graphical layout of all figures were subsequently manually edited and finalized by the authors using Canva. All figures were manually reviewed and verified to ensure consistency with the scientific evidence and the final manuscript. All decisions regarding the scientific content, methodology, literature selection, data extraction, interpretation, and conclusions were independently made and verified by the authors.